# Application Of ADNN For Background Subtraction In Smart Surveillance System

University Of Alberta
Department of Computing Science

Piyush Batra, Gagan Raj Singh, Neeraj Goyal

December 7, 2022

**Brief Abstract**

Object movement identification is one of the most researched problems in the field of computer vision. In this task, we try to classify a pixel as foreground or background. Even though numerous traditional machine learning and deep learning methods already exist for this problem, the two major issues with most of them are the need for large amounts of ground truth data and their inferior performance on unseen videos. Since every pixel of every frame has to be labeled, acquiring large amounts of data for these techniques gets rather expensive. Recently, Zhao et al. [1] proposed one of a kind Arithmetic Distribution Neural Network (ADNN) for universal background subtraction which utilizes probability information from the histogram of temporal pixels and achieves promising results. Building onto this work, we developed an intelligent video surveillance system that uses ADNN architecture for motion detection, trims the video with parts only containing motion, and performs anomaly detection on the trimmed video.

## 1 Literature review

Motion detection aims to find regions related to moving objects, and background subtraction is a widely used technique for this task. Herein, every pixel of each video frame is compared against their historical counterparts or a background model, depending on the technique, and then classified into foreground or background. Pixels that differ significantly from the reference are classified as moving objects or foreground, and static pixels are referred to as background. This section will discuss some of the previously proposed methods related to background subtraction, video surveillance, and anomaly detection.

Many background modeling techniques based on mathematical theories, like the temporal average [2], temporal median [3], or the histogram over time [4], have been proposed for motion detection by a stationary camera. But these are not robust to challenges in surveillance videos such as dynamic background, object shadow, camera jitter, weather conditions (either snow or rain), and variations in illumination. To overcome these problems, several motion-detection methods, like Temporal Differencing [5], Three-frame Difference [6], Gaussian mixture model [7],

DSTEI [9], etc, have been presented over the past years.

Temporal Differencing, proposed by Cheung et al. [5], is used for detecting temporal changes in intensity in video frames. However, its main drawback is that the detected objects are incomplete and poorly presented.

In the Gaussian mixture model proposed by Stauffer and Grimson [7], the temporal histogram of each pixel is modeled using a mixture of K Gaussian distributions to precisely model a dynamic background. This method produced a real-time tracker which can deal with lighting changes, repetitive motions from clutter, and long-term scene changes. Later, Chan et al. [8] proposed a Generalized Stauffer–Grimson (GSG) algorithm for background subtraction in dynamic scenes. In this method, the statistics required for online learning of dynamic texture are derived from generalizing the GMM proposed by Stauffer and Grimson [7].

The Difference-based Spatio-Temporal Entropy Image (DSTEI) by Jing et al. [9] is an entropy-based method for human motion detection. A Spatio-temporal histogram is generated by accumulated pixels obtained by the difference between consecutive images. This histogram is then normalized to calculate the degree of randomness and magnitude of entropy to denote the significance of motion. In this method, noises are assumed to follow Gaussian distribution. However, these assumptions, such as heavy shadows or sudden illumination changes, will be violated in some cases.

Soumyadip Sengupta et al. [10] proposed a background matting technique that generated high-quality foreground and alpha mattes in natural settings. In this method, a deep learning framework is developed and trained on synthetic-composite data and then adapted to actual data using an adversarial network. Even though providing an additional photo of the background requires a small amount of foresight, it is far less tedious than creating a trimap for traditional matting methods.

In 2017, Dan Yang et al. [11] proposed a multi-feature background approach for complex video scenes that measures the stability of features and then selects different dominant features to model the background from the pixel and time-sequence domains. This Stability of Adaptive Features approach showed promising results on both complex and baseline scenes.

For applications of background subtraction in real time, Z. Kuang et al. [12] proposed a combination of the Horn-Schunck optical-flow estimation technique [13] and autoencoder neural networks that solve the problem of motion blur in real-time background subtraction during video conferencing. This method uses an optical-flow-based model to extract motion features between every two frames and then combine these features with the appearance feature from the original frame. An encoder-decoder network in combination with CNN is then used to learn and predict a mask output for the human head and shoulders for background subtraction.

Similarly, for real-time background subtraction, DK Yadav et al. [14] proposed a Pixel Intensity Based (PIBBS) system that first models the background, then extracts moving objects with a threshold and updates the background using a feedback-based background updation scheme. To improve the detection quality, this system also uses morphological operators as the last step.

Bruno Sauvalle et al. [15] proposed using an autoencoder to model the background of a video as a low-dimensional manifold. The output of this autoencoder is then compared with the original image to compute the segmentation masks. In this method, the autoencoder is also trained to predict the background noise, which allows it to compute a pixel-dependent threshold for each frame to perform the foreground segmentation. Without using temporal or motion information, this method could perform at par with state-of-the-art solutions on CDnet 2014 [16] and LASIESTA [17] datasets.

To overcome the problem of camera jitter and sudden changes in illumination, Ye Tao et al. [18] proposed a generative architecture for unsupervised deep background modeling, which learns the parameters automatically and uses intensity and optical flow features between a reference and a target frame. This system generates a background with a probabilistic heat map of the color values for a given input frame. This method could also be applied to unseen videos without re-training. When tested, this method shows promising results over state-of-the-art [19][20][21] methods on the SBMnet dataset[22].

Guanfang Dong et al. [23] proposed a novel denoising neural network model called Feature-guided Denoising Convolutional Neural Network (FDCNN) to denoise the images produced by portable devices. This technique employed a hierarchical denoising framework driven by a feature masking layer. The feature extraction algorithm used in this method is based on Explainable Artificial Intelligence (XAI) for medical images. Similarly, Yingnan Ma et al. [24] proposed an Edge-guided Denoising Convolutional Neural Network which can preserve important edge information in ultrasound images when removing noise. This method increases the recognition of various organs in ultrasound images.

Jhony H. Giraldo et al. [25] proposed a new algorithm called Graph Background Subtraction (GraphBGS). It is composed of instance segmentation, background initialization, graph construction, and graph sampling. Unlike Deep Learning methods for background subtraction which require vast amounts of data, this method is a semi-supervised algorithm inspired by the theory of recovery of graph signals.

To generate descriptions of human actions and their interactions, Zijian Kuang et al. [26] proposed a technique that utilizes an Actor Relation Graph (ARG) based model with novel improvements for group activity recognition. This method also used MobileNet as the backbone to extract features from each video frame.

To accurately perform background subtraction in a freely moving camera, Zhao et al. [27] developed a novel method called "the integration of foreground and background cues." The underlying motivation in this technique is to utilize the exclusiveness between these cues to compensate for their corresponding defects. The foreground is segmented by combining superpixels with proximity under multiple levels.

As video resolution and, subsequently, the video size is increasing daily, Ruixing et al. [28] proposed a method to compute the optimal image resolution adaptively. This is achieved by exploiting the correlation between an image's gray-value distribution and resolution. This approach was proposed to increase the performance of multi-object online tracking and learning. A novel tracklet reliability assessment metric was also introduced in this paper to eliminate the incorrect samples and can recover occluded targets.

As a unique application of neural networks in multimedia, C. Sun et al. [29] proposed a 2-step product re-identification (Re-ID) method which involves image feature extraction and a feature search and retrieval engine. To extract the features of the input image, a novel AlphaAlexNet, an extended version of the AlexNet, is being used. Vearch, a visual search system, is used as the image search similarity engine. The new model - AlphaAlexNet, demonstrated improved object detection accuracy of Vearch.

To classify two distributions without using just histograms and incorporating a deep learning network to learn and classify distributions automatically, Chunqiu Zhao and Anup Basu [30] proposed a novel vessel segmentation method based on distribution learning using a spatial distribution descriptor (RPoSP) under multiple scales. Here, statistical distributions are indirectly forced as an input to a CNN for distribution learning. The proposed approach showed promising results when compared to existing state-of-the-art methods[31][32] on the DRIVE[33] dataset.

Yongxin Ge et al. [34] proposed the Deep Variation Transformation Network (DVTN) model, which uses pixel variations to detect the background. This model assigns the probability to each pixel, and then by using thresholding, it computes whether it's background or foreground. This model compares the pixel variation instead of distributions. Previously used models in background detection usually fail when they encounter similar observations, causing false detections. The DVTN analyzes the pixel variations in a new space, where the above observations are classified easily. This model outperforms the traditional background detection models by showing astonishing results on the CDnet2014 dataset.

However, all of the methods mentioned above require either a large amount of ground truth data or result in inadequate performance on unseen videos. Zhao and Basu [35] proposed a Deep Pixel Distribution Learning (DPDL) technique to overcome these issues. Unlike typical approaches, which compare new frames to a formulated background model, this technique focuses on comparing pixels' current and historical frames. This method uses a novel pixel-based feature called the Random Permutation of Temporal Pixels (RPoTP) to represent the distribution of past observations for a particular pixel. Subsequently, a CNN is used to learn whether the current pixel is foreground or background. Adding on to this method, Zhao et al. [36] later proposed a new Dynamic Deep Pixel Distribution Learning (D-DPDL) technique. In this method, the RPoTP feature is dynamically permuted in this method for every training epoch. To compensate for the random noise generated in this process, a Bayesian Refinement model is used and improve the accuracy.

Zhao et al. [1] also proposed an Arithmetic Distribution Neural Network architecture demonstrating even better performance than the D-DPDL method. The input in the ADNN method is histograms of subtractions between current pixels and their historical counterparts. The sum and product arithmetic distribution layers proposed here demonstrate a better ability to classify distributions than the convolutional layers in D-DPDL. Moreover, the number of learning parameters used in ADNN architecture (0.1 Million) is significantly less than that used in the D-DPDL method (7 Million).

Coming onto detecting anomalies in videos, Virender Singh et al. [37] proposed an approach to detect variation from the norm in real-world CCTV recordings. This method uses two deep learning models (CNN and RNN) to learn a general anomaly detection model with a poorly labeled dataset. The training dataset has been doubled by flipping the videos horizontally, thus increasing the testing accuracy. The overall accuracy of the model is 97.23

Y Fan et al. [38] proposed a technique that first converts the video clips of an ongoing event into Dynamic Images, which can simultaneously capture the appearance and temporal evolution of the occurrence. The approach uses dynamic images of two categories of video clips and involves training a detector based on deep-learning techniques.

Yu Tian et al. [39] proposed a weakly-supervised anomaly detection algorithm, Robust Temporal Feature Magnitude learning (RTFM), aiming to identify snippets containing abnormal events. This method trains a feature magnitude learning function to effectively recognize the positive instances, substantially enhancing the robustness of this method to the negative instances from abnormal videos. RTFM achieves significantly improved subtle anomaly discriminability and sample efficiency.

The Weakly Supervised Video Anomaly Detection(WSVAD) [40]-[42] method for anomaly detection suffers from the wrong identification of normal and abnormal instances during the training process. Kapil Deshpande et al. [43] proposed better-quality transformer-based features named Videoswin Features, followed by an attention layer to capture long and short-range dependencies in the temporal domain. This method extracts better-quality features from available videos resulting in better performance.

## 2 Method

In this work, we implemented an Arithmetic Distribution Neural Network [1] to develop a video surveillance system for identifying object movement in a static video. In this ADNN model, the arithmetic operations are utilized to introduce the arithmetic distribution layers, including the product and sum distribution layers. Outputs from these layers are combined and passed through a classifier for accurate classification. We chose this architecture because it requires training only one network, with limited training data, and it works well with unseen test videos.

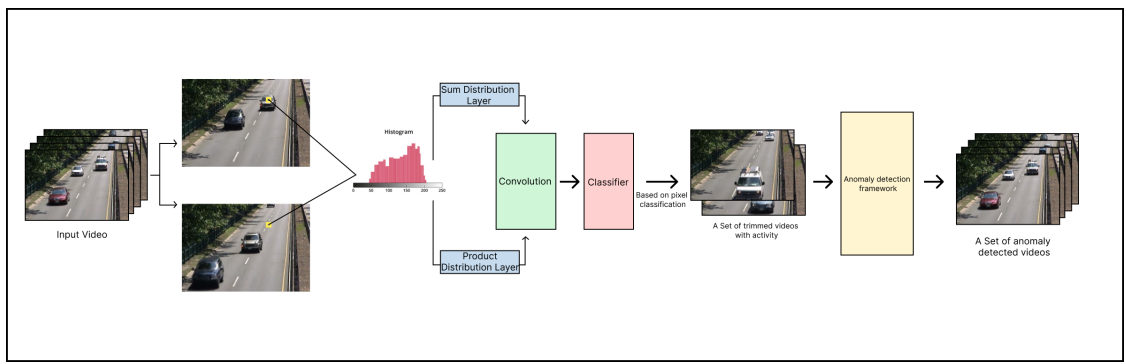

*Figure 1:* The flow diagram of our proposed approach

Upon successful object movement detection using background subtraction, we further analyzed the results obtained from ADNN to filter out their anomalous activities.

### 2.1 Motion detection - Arithmetic Distribution Neural Network

In this work, we used ADNN proposed by Zhao et al. [1] to detect motion in the input surveillance video. This paper proposed arithmetic distribution layers, which are a new type of network layer that is designed to improve distribution analysis in classification tasks. These layers, which include product and sum distribution layers, are an alternative to convolution layers. During the forward pass of the proposed arithmetic distribution layers, the input distributions are processed using the distributions in the learning kernels to generate the output distributions. In the backpropagation process, the gradient of the distributions in the learning kernels with respect to the network output is calculated to update the learning kernels. These operations are based on histograms and arithmetic distribution operations rather than the matrix arithmetic operations used in traditional convolution layers.

To improve the accuracy of the foreground mask generated, an improved Bayesian refinement model is used. This model takes into account the correlations between pixels by using a mixture of Gaussian approximation functions rather than just Euclidean distance, as in the original Bayesian refinement model. The Bayesian refinement model is used to iteratively refine the foreground mask, with the output of the arithmetic distribution neural network serving as the initial binary mask for the iteration process.

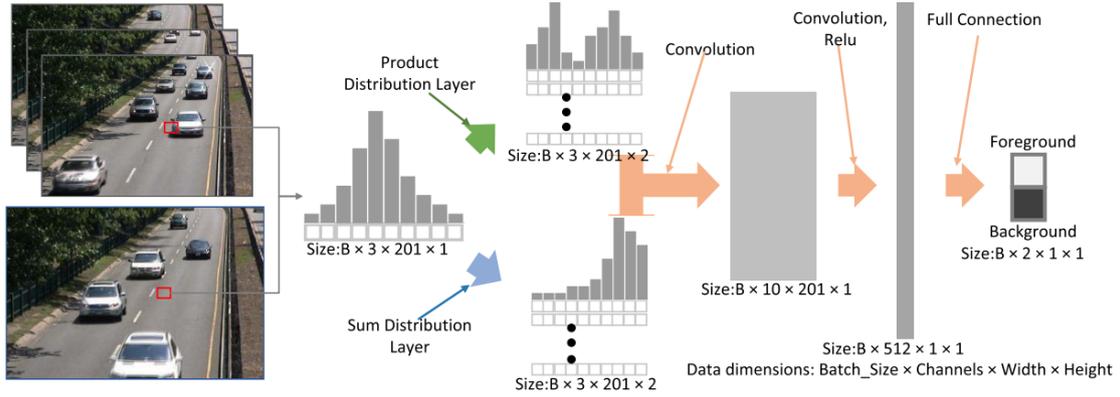

*Figure 2:* Arithmetic distribution neural network for background subtraction

After obtaining the refined foreground masks from the ADNN architecture, we utilize a python script to generate a trimmed video from a set of input frames by using a threshold value on the frames generated by the Bayesian refinement model. The threshold value determines the minimum number of white pixels (foreground pixels) that must be present in a frame in order for it to be included in the trimmed video. For this work, we are using a threshold value of 5% to generate the trimmed videos.

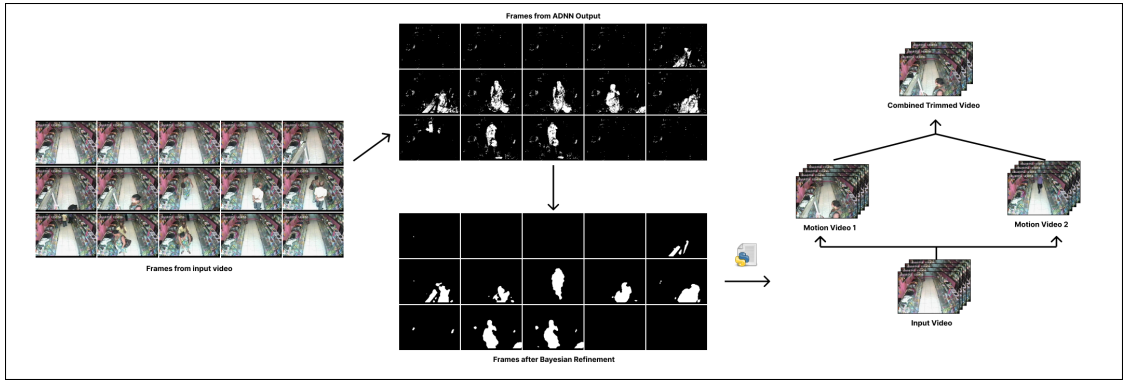

*Figure 3:* Generation of trimmed video from input frames passed through the ADNN (arithmetic distribution neural network)

## 2.2 Anomaly Detection

Following the works of Waqas Sultani et al. [44], we have put into use their novel Multiple Instance Learning framework for the second part of our system. Once we obtain the trimmed video from the previous step, we use that as the input in this step. In this, a training set of positive (containing an abnormality someplace) and negative (having no anomaly) videos are used to train the anomaly detection model. Then each video is divided into a sequence of non-overlapping temporal segments.

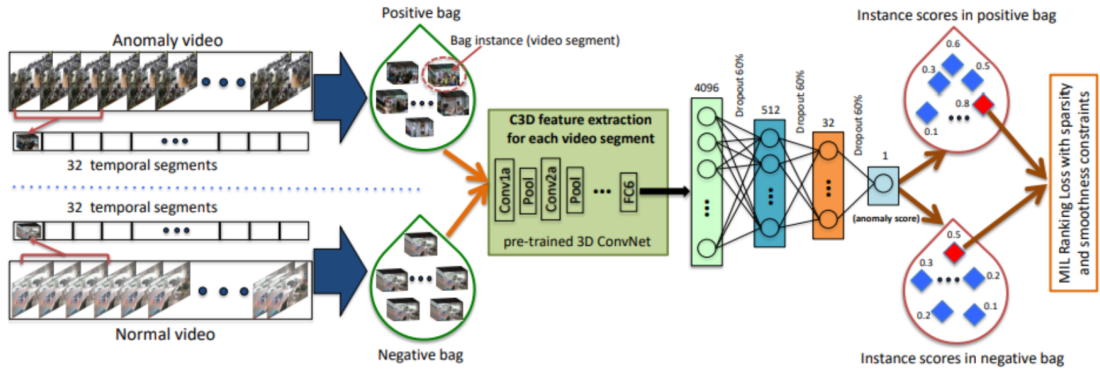

*Figure 4:* Anomaly detection flow diagram

Each video in the training set can be represented as a bag, and each video segment represents an instance in the bag. After extracting C3D features from video segments using a pre-trained 3D convNet, a fully connected neural network is trained using the novel ranking loss function; it computes the ranking loss between the top-rated occurrences in the positive bag and the negative bag.

In conclusion, the proposed method for detecting anomalies in surveillance videos consists of two main steps. First, the ADNN architecture is used to detect motion in the input video and generate a refined foreground mask. This mask is then used to create a trimmed video, which is used as input for the second step of the system. In this step, we used a pre-trained multiple instance learning model trained on a set of positive and negative videos and used to classify each temporal segment in the test video as normal or anomalous. The predicted scores for each segment are then combined to generate a prediction (anomaly graph) for the entire video. By combining these two approaches, the system is able to effectively detect abnormalities in surveillance videos, even when they only occur for a short period of time or are only present in a small number of segments.

## 3  Results

In this section, we will discuss our experimental results for two different videos. Table 1 compares the full video and trimmed video for two different videos, labeled Video 1 and Video 2. For Video 1, the full video had a duration of 06:37 minutes, a size of 90.5 MB, and contained 11937 frames. The anomaly detection process for this video took 789 seconds. The trimmed video for Video 2 had a duration of 04:09 minutes, a size of 68.5 MB, and contained 7470 frames. The anomaly detection process for this video took 540 seconds, which is lower than the time taken for the full video.

For Video 2, the full video had a duration of 04:59 minutes, a size of 40.6 MB, and contained 8990 frames. The anomaly detection process for this video took 610 seconds. On the other hand, the trimmed video for Video 2 had a duration of 1:04 minutes, a size of 10.2 MB, and contained 1950 frames. The anomaly detection process for this video took 137 seconds, which is also lower than the time taken for the full video.

|  |  | Duration (mm: ss) | Size (MB) | Frames | Anomaly Detection (cpu - sec) |
|---|---|---|---|---|---|
| Video 1 | Full Video | 06:37 | 90.5 | 11937 | 789 |
|  | Trimmed Video (Combined) | 04:09 | 68.5 | 7470 | 540 |
| Video2 | Full Video | 04:59 | 40.6 | 8990 | 610 |
|  | Trimmed Video (Combined) | 1:04 | 10.2 | 1950 | 137 |

*Table 1: Comparison of results for trimmed and full-length videos*

The graphs obtained after anomaly detection are shown below. These are the relative anomaly scores of each video segment (32 in this case). We can see that the anomalous regions in the trimmed video are more focused, and there are comparatively fewer inactive regions. Moreover, the overall structure of the graphs is similar for both the original and trimmed videos, indicating that trimming down the video does not affect the anomaly identification and the relative scores of different segments.

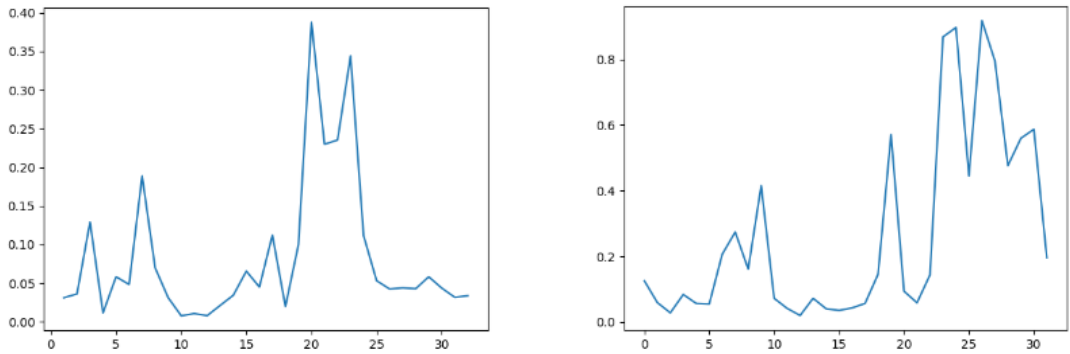

*Figure 5:* The graphs indicate anomaly scores of the video2 (left) and its trimmed version (right)

Overall, the results in Table 1 show that the trimmed videos had shorter durations and smaller sizes compared to the full videos. Additionally, the anomaly detection process for the trimmed videos took much less time than the full videos in both examples. This suggests that using trimmed videos leads to a more efficient anomaly detection process.

## 4 Discussion

The results presented above demonstrate the effectiveness of our ADNN-based video surveillance system in identifying object movement and filtering out anomalous activities. As shown, the trimmed videos had shorter durations, smaller sizes and required less time for anomaly detection compared to the full videos in both examples. This suggests that the ADNN model and the use of trimmed videos lead to a more efficient and effective video surveillance system. Additionally, the ADNN model we employed has the advantage of requiring only limited training data and being able to perform well with unseen test videos. This makes it a suitable choice for practical implementation in real-world scenarios.

In conclusion, our ADNN-based video surveillance system has demonstrated its ability to accurately detect object movement and filter out anomalous activities, making it a promising solution for video surveillance applications.

## 5  Future Work

In the future, we plan to work on making the ADNN model more efficient at inferring foreground masks, as it currently takes a significant amount of time to process videos. This will be a major challenge, but we believe it is necessary in order to make the system more practical and useful in real-world scenarios.

Additionally, we will work on generating a better test dataset to further evaluate the adaptability of this system. This will help us to better understand the limitations and potential improvements of the system. Overall, the goal would be to improve the efficiency of the ADNN model in order to make it a useful tool for video surveillance and anomaly detection applications.

## References

[1] Zhao, C., Hu, K., Basu, A. (2022). Universal Background Subtraction Based on Arithmetic Distribution Neural Network. IEEE Transactions on Image Processing, 31, 2934–2949. https://doi.org/10.1109/tip.2022.3162961

[2] Lee, B.; Hedley, M. Background estimation for video surveillance. In: Image Vision Computing New Zealand (IVCNZ '02); 2002; Auckland, NZ. 2002. 315-320. http://hdl.handle.net/102.100.100/199802

[3] Graszka, P. (2014). Median mixture model for background–foreground segmentation in video sequences.

[4] Roy, S. M., Ghosh, A. (2018). Real-Time Adaptive Histogram Min-Max Bucket (HMMB) Model for Background Subtraction. IEEE Transactions on Circuits and Systems for Video Technology, 28(7), 1513–1525. https://doi.org/10.1109/tcsvt.2017.2669362

[5] Cheung, S.-C. S., Kamath, C. (2005). Robust Background Subtraction with Foreground Validation for Urban Traffic Video. EURASIP Journal on Advances in Signal Processing, 2005(14). https://doi.org/10.1155/asp.2005.2330

[6] Chen, C., Zhang, X. (2012). Moving Vehicle Detection Based on Union of Three-Frame Difference. Lecture Notes in Electrical Engineering, 459–464. https://doi.org/10.1007/978-3-642-27296-7_71

[7] Stauffer, C., Grimson, W. E. L. (2022). Adaptive background mixture models for real-time tracking. Proceedings. 1999 IEEE Computer Society Conference on Computer Vision and Pattern Recognition (Cat. No PR00149). https://doi.org/10.1109/cvpr.1999.784637

[8] Chan, A. B., Mahadevan, V., Vasconcelos, N. (2010). Generalized Stauffer–Grimson background subtraction for dynamic scenes. Machine Vision and Applications, 22(5), 751–766. https://doi.org/10.1007/s00138-010-0262-3

[9] Guo-Jing, Chng Eng Siong, Rajan, D. (2022). Foregroung motion detection by difference-based spatial temporal entropy image. 2004 IEEE Region 10 Conference TENCON 2004. https://doi.org/10.1109/tencon.2004.1414436

[10] Sengupta, S., Jayaram, V., Curless, B., Seitz, S. M., Kemelmacher-Shlizerman, I. (2020). Background Matting: The World Is Your Green Screen. 2020 IEEE/CVF Conference on Computer Vision and Pattern Recognition (CVPR). https://doi.org/10.1109/cvpr42600.2020.00236

[11] Yang, D., Zhao, C., Zhang, X., Huang, S. (2018). Background Modeling by Stability of Adaptive Features in Complex Scenes. IEEE Transactions on Image Processing, 27(3), 1112–1125. https://doi.org/10.1109/tip.2017.2768828

[12] Kuang, Z., Tie, X. (2021). Flow-based Video Segmentation for Human Head and Shoulders. ArXiv.org. https://doi.org/10.48550/arXiv.2104.09752

[13] sniklaus. (2022, March 16). sniklaus/pytorch-pwc: a reimplementation of PWC-Net in PyTorch that matches the official Caffe version. GitHub. https://github.com/sniklaus/pytorch-pwc

[14] Yadav, D. K., Sharma, L., Bharti, S. K. (2014). Moving object detection in real-time visual surveillance using background subtraction technique. 2014 14th International Conference on Hybrid Intelligent Systems. https://doi.org/10.1109/his.2014.7086176

[15] Sauvalle, B., de. (2021). Autoencoder-based background reconstruction and foreground segmentation with background noise estimation. ArXiv.org. https://doi.org/10.48550/arXiv.2112.08001

[16] Wang, Y., Jodoin, P.-M., Porikli, F., Konrad, J., Benezeth, Y., Ishwar, P. (2014). CDnet 2014: An Expanded Change Detection Benchmark Dataset. 2014 IEEE Conference on Computer Vision and Pattern Recognition Workshops. https://doi.org/10.1109/cvprw.2014.126

[17] Cuevas, C., Yáñez, E. M., García, N. (2016). Labeled dataset for integral evaluation of moving object detection algorithms: LASIESTA. Computer Vision and Image Understanding, 152, 103–117. https://doi.org/10.1016/j.cviu.2016.08.005

[18] Tao, Y., Palasek, P., Ling, Z., Patras, I. (2017). Background modelling based on generative unet. 2017 14th IEEE International Conference on Advanced Video and Signal Based Surveillance (AVSS). https://doi.org/10.1109/avss.2017.8078483

[19] De Gregorio, M., Giordano, M. (2017). Background estimation by weightless neural networks. Pattern Recognition Letters, 96, 55–65. https://doi.org/10.1016/j.patrec.2017.05.029

[20] Ramirez-Alonso, G., Ramirez-Quintana, J. A., Chacon-Murguia, M. I. (2017). Temporal weighted learning model for background estimation with an automatic re-initialization stage and adaptive parameters update. Pattern Recognition Letters, 96, 34–44. https://doi.org/10.1016/j.patrec.2017.01.011

[21] Laugraud, B., Pierard, S., Van Droogenbroeck, M. (2016). LaBGen-P: A pixel-level stationary background generation method based on LaBGen. 2016 23rd International Conference on Pattern Recognition (ICPR). https://doi.org/10.1109/icpr.2016.7899617

[22] Jodoin, P.-M., Maddalena, L., Petrosino, A., Wang, Y. (2017). Extensive Benchmark and Survey of Modeling Methods for Scene Background Initialization. IEEE Transactions on Image Processing, 26(11), 5244–5256. https://doi.org/10.1109/tip.2017.2728181

[23] Dong, G., Ma, Y., Basu, A. (2021). Feature-Guided CNN for Denoising Images From Portable Ultrasound Devices. IEEE Access, 9, 28272–28281. https://doi.org/10.1109/access.2021.3059003

[24] Ma, Y., Yang, F., Basu, A. (2021). Edge-guided CNN for Denoising Images from Portable Ultrasound Devices. 2020 25th International Conference on Pattern Recognition (ICPR). https://doi.org/10.1109/icpr48806.2021.9412758

[25] Giraldo, J. H., Bouwmans, T. (2020). GraphBGS: Background Subtraction via Recovery of Graph Signals. ArXiv.org. https://doi.org/10.48550/arXiv.2001.06404

[26] Kuang, Z., Tie, X. (2020). Improved Actor Relation Graph based Group Activity Recognition. ArXiv.org. https://doi.org/10.48550/arXiv.2010.12968

[27] Zhao, C., Sain, A., Qu, Y., Ge, Y., Hu, H. (2019). Background Subtraction Based on Integration of Alternative Cues in Freely Moving Camera. IEEE Transactions on Circuits and Systems for Video Technology, 29(7), 1933–1945. https://doi.org/10.1109/tcsvt.2018.2854273

[28] Yu, R., Cheng, I., Zhu, B., Bedmutha, S., Basu, A. (2018). Adaptive Resolution Optimization and Tracklet Reliability Assessment for Efficient Multi-Object Tracking. IEEE Transactions on Circuits and Systems for Video Technology, 28(7), 1623–1633. https://doi.org/10.1109/tcsvt.2017.2668278

[29] Sun, C., Song, L. B. (2021). Product Re-identification System in Fully Automated Defect Detection. ArXiv.org. https://doi.org/10.48550/arXiv.2112.10324

[30] Zhao, C., Basu, A. (2021). Pixel Distribution Learning for Vessel Segmentation under Multiple Scales. 2021 43rd Annual International Conference of the IEEE Engineering in Medicine Biology Society (EMBC). https://doi.org/10.1109/embc46164.2021.9629614

[31] Fu, H., Xu, Y., Lin, S., Kee Wong, D. W., Liu, J. (2016). DeepVessel: Retinal Vessel Segmentation via Deep Learning and Conditional Random Field. Medical Image Computing and Computer-Assisted Intervention – MICCAI 2016, 132–139. https://doi.org/10.1007/978-3-319-46723-8_16

[32] Dasgupta, A., Singh, S. (2017). A fully convolutional neural network based structured prediction approach towards the retinal vessel segmentation. 2017 IEEE 14th International Symposium on Biomedical Imaging (ISBI 2017). https://doi.org/10.1109/isbi.2017.7950512

[33] Staal, J., Abramoff, M. D., Niemeijer, M., Viergever, M. A., van Ginneken, B. (2004). Ridge-Based Vessel Segmentation in Color Images of the Retina. IEEE Transactions on Medical Imaging, 23(4), 501–509. https://doi.org/10.1109/tmi.2004.825627

[34] Ge, Y., Zhang, J., Ren, X., Zhao, C., Yang, J., Basu, A. (2021). Deep Variation Transformation Network for Foreground Detection. IEEE Transactions on Circuits and Systems for Video Technology, 31(9), 3544–3558. https://doi.org/10.1109/tcsvt.2020.3042559

[35] Zhao, C., Cham, T.-L., Ren, X., Cai, J., Zhu, H. (2018). Background Subtraction Based on Deep Pixel Distribution Learning. 2018 IEEE International Conference on Multimedia and Expo (ICME). https://doi.org/10.1109/icme.2018.8486510

[36] Zhao, C., Basu, A. (2020). Dynamic Deep Pixel Distribution Learning for Background Subtraction. IEEE Transactions on Circuits and Systems for Video Technology, 30(11), 4192–4206. https://doi.org/10.1109/tcsvt.2019.2951778

[37] Singh, V., Singh, S., Gupta, P. (2020). Real-Time Anomaly Recognition Through CCTV Using Neural Networks. Procedia Computer Science, 173, 254–263. https://doi.org/10.1016/j.procs.2020.06.030

[38] Fan, Y., Wen, G., Li, D., Qiu, S., Levine, M. D. (2018). Early event detection based on dynamic images of surveillance videos. Journal of Visual Communication and Image Representation, 51, 70–75. https://doi.org/10.1016/j.jvcir.2018.01.002

[39] Tian, Y., Pang, G., Chen, Y., Singh, R., Verjans, Johan W, Carneiro, G. (2021). Weakly-supervised Video Anomaly Detection with Robust Temporal Feature Magnitude Learning. ArXiv.org. https://doi.org/10.48550/arXiv.2101.10030

[40] Liu, W., Luo, W., Li, Z., Zhao, P., Gao, S. (2019). Margin Learning Embedded Prediction for Video Anomaly Detection with A Few Anomalies. Ijcai.org, 3023–3030. https://www.ijcai.org/proceedings/2019/419

[41] Pang, G., Cao, L., Chen, L., Liu, H. (2018). Learning Representations of Ultrahigh-dimensional Data for Random Distance-based Outlier Detection. Proceedings of the 24th ACM SIGKDD International Conference on Knowledge Discovery Data Mining. https://doi.org/10.1145/3219819.3220042

[42] Pang, G., Shen, C., Anton. (2019). Deep Anomaly Detection with Deviation Networks. ArXiv.org. https://doi.org/10.48550/arXiv.1911.08623

[43] Deshpande, K., Punn, N. S., Sonbhadra, Sanjay Kumar, Agarwal, S. (2022). Anomaly detection in surveillance videos using transformer-based attention model. ArXiv.org. https://doi.org/10.48550/arXiv.2206.01524

[44] Sultani, W., Chen, C., Shah, M. (2018). Real-world Anomaly Detection in Surveillance Videos. ArXiv.org. https://doi.org/10.48550/arXiv.1801.04264

[45] zhaochenqiu. (2022, June 6). zhaochenqiu/UBgS_ADNNet. GitHub. https://github.com/zhaochenqiu/UBgS_ADNNet